\documentclass[letterpaper, 10 pt, journal, twoside]{IEEEtran}
\pdfoutput=1
\IEEEoverridecommandlockouts                              
\usepackage{hyperref}
\hypersetup{hidelinks,pdfnewwindow,breaklinks} 
\usepackage{mathptmx}
\usepackage{times}
\usepackage{cite}
\usepackage{amsmath,amssymb,amsfonts}
\usepackage[ruled,vlined]{algorithm2e}
\usepackage{graphicx}
\usepackage{textcomp}
\usepackage{bm}
\usepackage{listings}
\usepackage[caption=false,font=normalsize,labelfont=sf,textfont=sf]{subfig}
\usepackage{mathtools}
\usepackage{color}
\usepackage{import}
\usepackage{mathrsfs}
\usepackage{boxedminipage}
\definecolor{caltechblue}{rgb}{0,0.2314,0.5}
\definecolor{caltechgreen}{rgb}{0,0.3451,0.3137}
\definecolor{caltechorange}{rgb}{1,0.4235,0.0406}
\definecolor{caltechred}{rgb}{0.4805,0.1875,0.2422}

\makeatletter
\def\ps@IEEEtitlepagestyle{
\def\@oddhead{\hbox{}\@IEEEheaderstyle\leftmark\hfil\thepage}\relax
\def\@evenhead{\@IEEEheaderstyle\thepage\hfil\leftmark\hbox{}}\relax
  \def\@oddfoot{\mycopyrightnotice}
  \def\@evenfoot{}
}

\def\mycopyrightnotice{
  {\footnotesize
  \begin{boxedminipage}{\textwidth}
  \centering
  © 2020 IEEE. Personal use of this material is permitted. Permission from IEEE must be obtained for all other uses, in any current or future media, including reprinting/republishing this material for advertising or promotional purposes, creating new collective works, for resale or redistribution to servers or lists, or reuse of any copyrighted component of this work in other works. Digital Object Identifier (DOI): {\color{caltechgreen}\underline{\href{https://ieeexplore.ieee.org/document/9302618}{10.1109/LCSYS.2020.3046529}}}
  \end{boxedminipage}
  }
}

\def\BibTeX{{\rm B\kern-.05em{\sc i\kern-.025em b}\kern-.08em
    T\kern-.1667em\lower.7ex\hbox{E}\kern-.125emX}}
\newtheorem{theorem}{Theorem}
\newtheorem{lemma}{Lemma}

\newtheorem{proposition}{Proposition}
\newtheorem{remark}{Remark}

\newtheorem{corollary}{Corollary}
\newtheorem{example}{Example}

\newcommand{\ie}{{i}.{e}.}
\newcommand{\eg}{{e}.{g}.}
\newcommand{\st}{{s}.{t}.}

\DeclareMathOperator{\sym}{sym}

\title{Neural Stochastic Contraction Metrics for Learning-based Control and Estimation}

\author{Hiroyasu Tsukamoto\IEEEauthorrefmark{1},  Soon-Jo Chung\IEEEauthorrefmark{1}, and Jean-Jacques E. Slotine\IEEEauthorrefmark{2}
\thanks{\IEEEauthorrefmark{1} Graduate Aerospace Laboratories, California Institute of Technology, Pasadena, CA, {{\tt\small\{htsukamoto, sjchung\}@caltech.edu}}.}
\thanks{\IEEEauthorrefmark{2} Nonlinear Systems Laboratory, Massachusetts Institute of Technology,
Cambridge MA, {{\tt\small jjs@mit.edu}}.}
\thanks{Code: {\color{caltechgreen}\underline{\href{https://github.com/astrohiro/nscm}{https://github.com/astrohiro/nscm}}}.}
}

\begin{document}
\maketitle
\begin{abstract}
We present Neural Stochastic Contraction Metrics (NSCM), a new design framework for provably-stable robust control and estimation for a class of stochastic nonlinear systems. It uses a spectrally-normalized deep neural network to construct a contraction metric, sampled via simplified convex optimization in the stochastic setting. Spectral normalization constrains the state-derivatives of the metric to be Lipschitz continuous, thereby ensuring exponential boundedness of the mean squared distance of system trajectories under stochastic disturbances. The NSCM framework allows autonomous agents to approximate optimal stable control and estimation policies in real-time, and outperforms existing nonlinear control and estimation techniques including the state-dependent Riccati equation, iterative LQR, EKF, and the deterministic neural contraction metric, as illustrated in simulation results.
\end{abstract}
\begin{IEEEkeywords}
Machine learning, Stochastic optimal control, Observers for nonlinear systems.
\end{IEEEkeywords}
\section{Introduction}
\label{introduction}
The key challenge for control and estimation of autonomous aerospace and robotic systems is how to ensure optimality and stability. Oftentimes, their motions are expressed as nonlinear systems with unbounded stochastic disturbances, the time evolution of which is expressed as It\^{o} stochastic differential equations~\cite{nla.cat-vn712853}. As their onboard computational power is often limited, it is desirable to execute control and estimation policies computationally as cheaply as possible.

\begin{figure}
    \centering
    \includegraphics[width=85mm]{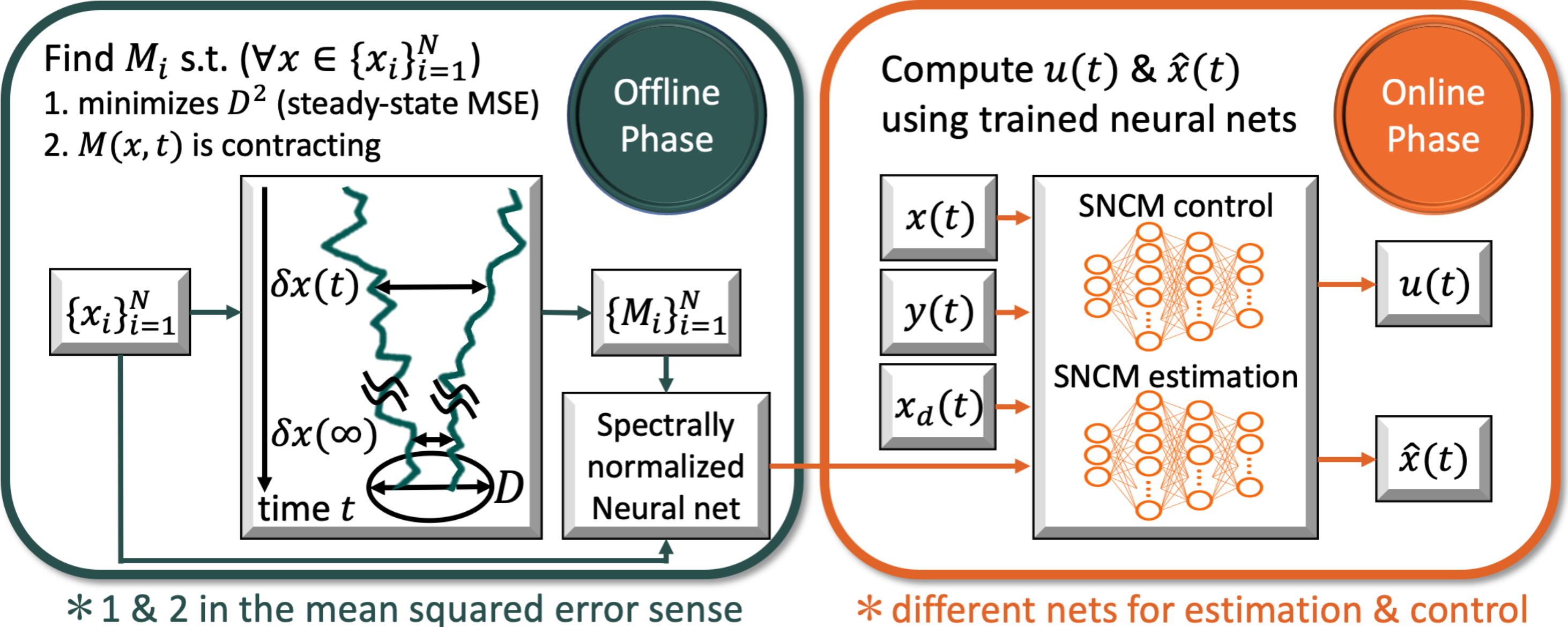}
    \caption{Illustration of NSCM ($M(x,t)$: optimal contraction metric; $x_i$ and $M_i$: sampled states and contraction metrics; $y(t)$: measurements; $x(t)$, $x_d(t)$, and $\hat{x}(t)$: actual, target, and estimated trajectories, respectively.}
    \label{nscmdrawing}
\end{figure}

In this paper, we present a Neural Stochastic Contraction Metric (NSCM) based robust control and estimation framework outlined in Fig.~\ref{nscmdrawing}.
It uses a spectrally-normalized neural network as a model for an optimal contraction metric (differential Lyapunov function), the existence of which guarantees exponential boundedness of the mean squared distance between two system trajectories perturbed by stochastic disturbances. Unlike the Neural Contraction Metric (NCM)~\cite{ncm}, where we proposed a learning-based construction of optimal contraction metrics for control and estimation of nonlinear systems with bounded disturbances, stochastic contraction theory~\cite{4806161,scrm,observer} guarantees stability and optimality in the mean squared error sense for unbounded stochastic disturbances via convex optimization. Spectral Normalization (SN)~\cite{miyato2018spectral} is introduced in the NSCM training, in order to validate a major assumption in stochastic contraction that the first state-derivatives of the metric are Lipschitz. We also extend the State-Dependent-Coefficient (SDC) technique~\cite{sddre} further to include a target trajectory in control and estimation, for the sake of global exponential stability of unperturbed systems.

In the offline phase, we sample contraction metrics by solving convex optimization to minimize an upper bound of the steady-state mean squared distance of stochastically perturbed system trajectories (see Fig.~\ref{nscmdrawing}). Other convex objectives such as control effort could be used depending on the application of interest. We call this method the modified CV-STEM (mCV-STEM), which differs from the original work~\cite{mypaperTAC} in the following points: 1) a simpler stochastic contraction condition with an affine objective function both in control and estimation, thanks to the Lipschitz condition on the first derivatives of the metrics; 2) generalized SDC parameterization, \ie{}, $A$ \st{} $A(x,x_d,t)(x-x_d)=f(x,t)+B(x,t)u_d-f(x_d,t)-B(x_d,t)u_d$ instead of $A(x,t)x=f(x,t)$, for systems $\dot{x} = f(x,t)+B(x,t)u$, which results in global exponential stability of unperturbed systems even with a target trajectory, $x_d$ for control and $x$ for estimation; and 3) optimality in the contraction rate $\alpha$ and disturbance attenuation parameter $\varepsilon$.
The second point is in fact general, since $A$ can always be selected based on the line integral of the Jacobian of $f(x,t)+B(x,t)u_d$, a property which can also be applied to the deterministic NCM setting of~\cite{ncm}. We then train a neural network with the sampled metrics subject to the aforementioned Lipschitz constraint using the SN technique. Note that reference-independent integral forms of control laws~\cite{7852456,7989693,47710,WANG201944,vccm} could be considered by changing how we sample the metrics in this phase. Our contraction-based formulation enables larger contracting systems to be
built recursively by exploiting combination properties~\cite{contraction}, as in systems with hierarchical combinations (\eg{} output feedback or negative feedback), or to consider systems with time-delayed communications~\cite{1618853}.

In the online phase, the trained NSCM models are exploited to approximate the optimal control and estimation policies, which only require one neural network evaluation at each time step as shown in Fig~\ref{nscmdrawing}. The benefits of this framework are demonstrated in the rocket state estimation and control problem, by comparing it with the State-Dependent Riccati Equation (SDRE) method~\cite{sddre,observer}, Iterative LQR (ILQR)~\cite{ilqr,silqr}, EKF, NCM, and mCV-STEM.
\subsubsection*{Related Work}
Contraction theory~\cite{contraction} is an analytical tool for studying the differential dynamics of a nonlinear system under a contraction metric, whose existence leads to a necessary and sufficient characterization of its exponential incremental stability. The theoretical foundation of this paper rests on its extension to stability analysis of stochastic nonlinear systems~\cite{4806161,scrm,observer}. The major difficulty in applying it in practice is the lack of general analytical schemes to obtain a suitable stochastic contraction metric for nonlinear systems written as It\^{o} stochastic differential equations~\cite{nla.cat-vn712853}.

For deterministic systems, there are several learning-based techniques for designing real-time computable optimal Lyapunov functions/contraction metrics. These include~\cite{ncm,spencer18lyapunovnn,NIPS2019_8587}, where neural networks are used to represent the optimal solutions to the problem of obtaining a Lyapunov function. This paper improves our deterministic NCM~\cite{ncm}, as the NSCM explicitly considers the case of stochastic nonlinear systems, where deterministic control and estimation policies could fail due to additional derivative terms in the differential of the contraction metric under stochastic perturbation.

The CV-STEM~\cite{mypaperTAC} is derived to construct a contraction metric accounting for the stochasticity in dynamical processes. It is designed to minimize the upper bound of the steady-state mean squared tracking error of stochastic nonlinear systems, assuming that the first and second derivatives of the metric with respect to its state are bounded. In this paper, we only assume that the first derivatives are Lipschitz continuous, thereby enabling the use of spectrally-normalized neural networks~\cite{miyato2018spectral}. This also significantly reduces the computational burden in solving the CV-STEM optimization problems, allowing autonomous agents to perform both optimal control and estimation tasks in real-time.
\section{Preliminaries}
\label{preliminaries}
We use $\|x\|$ and $\|A\|$ for the Euclidean and induced 2-norm, $I$ for the identity matrix, $E[\cdot]$ for the expected value, $\sym(A) = (A+A^T)/2$, and $A \succ 0$, $A \succeq 0$, $A \prec 0$, and $A \preceq 0$ for positive definite, positive semi-definite, negative definite, and negative semi-definite matrices, respectively. Also, $f_{x}$ is the partial derivative of $f(x,t)$ respect to the state $x$, and $M_{x_i}$ is of $M(x,t)$ with respect to the $i$th element of $x$, $M_{x_ix_j}$ is of $M(x,t)$ with respect to the $i$th and $j$th elements of $x$.
\subsection{Neural Network and Spectral Normalization}
\label{spectral}
A neural network is a mathematical model for representing training samples $\{(x_i,y_i)\}_{i=1}^{N}$ of $y = \phi(x)$ by optimally tuning its hyperparameters $W_{\ell}$, and is given as
\begin{align}
\label{neuralnet}
y_i = \varphi(x_i;W_{\ell}) = T_{L+1}*\sigma*T_{L}*\cdots*\sigma*T_{1}(x_i)
\end{align}
where $T_{\ell}(x) = W_{\ell}x$, $*$ denotes composition of functions, and $\sigma$ is an activation function $\sigma(x) = \tanh(x)$. Note that $\varphi(x) \in C^{\infty}$. 

Spectral normalization (SN)~\cite{miyato2018spectral} is a technique to overcome the instability of neural network training by constraining (\ref{neuralnet}) to be globally Lipschitz, \ie{},  $\exists \ L_{nn} \ge 0$ \st{} $\|\varphi(x)-\varphi(x')\| \leq L_{nn}\|x-x'\|,~\forall x,x'$, which is shown to be useful in nonlinear control designs~\cite{8794351}.
SN normalizes the weight matrices $W_{\ell}$ as $W_{\ell} = (C_{nn}\Omega_{\ell})/\|\Omega_{\ell}\|$ with $C_{nn} \geq 0$ being a given constant, and trains a network with respect to $\Omega_{\ell}$. Since this results in ${\|\varphi(x)-\varphi(x')\|} \leq C_{nn}^{L+1}\ \|x-x'\| $~\cite{miyato2018spectral}, setting $C_{nn}=L_{nn}^{1/(L+1)}$ guarantees Lipschitz continuity of $\varphi(x)$. In Sec.~\ref{sec_sn_nscm}, we propose one way to use SN for building a neural network that guarantees the Lipschitz assumption on $M_{x_i}$ in Theorem~\ref{sic}.
\subsection{Stochastic Contraction Analysis for Incremental Stability}
Consider the following nonlinear system with stochastic perturbation given by the It\^{o} stochastic differential equation:
\begin{align}
\label{stochastic_dynamics}
dx =& f(x,t)dt + G(x,t)d\mathcal{W}(t),~x(0) = x_0
\end{align}
where $t\in \mathbb{R}_{\geq0}$, $x:\mathbb{R}_{\geq0} \to \mathbb{R}^{n}$, $f:\mathbb{R}^n\times\mathbb{R}_{\geq0} \to \mathbb{R}^{n}$, $G:\mathbb{R}^n\times\mathbb{R}_{\geq0} \to \mathbb{R}^{n\times d}$, $\mathcal{W}(t)$ is a $d$-dimensional Wiener process, and $x_0$ is a random variable independent of $\mathcal{W}(t)$~\cite{arnold_SDE}. We assume that 1)~$\exists L_1 > 0$ \st{} $\|f(x_1,t)-f(x_2,t)\|+\|G(x_1,t)-G(x_2,t)\|_F \leq L_1 \|x_1-x_2\|$, $\forall t\in\mathbb{R}_{\geq0}$ and $\forall x_1,~x_2 \in \mathbb{R}^n$, and 2)~$\exists L_2 > 0$, \st{} $\|f(x_1,t)\|^2+\|G(x_1,t)\|_F^2 \leq L_2 (1+\|x_1\|^2)$, $\forall t\in\mathbb{R}_{\geq0}$ and $\forall x_1 \in \mathbb{R}^n$ for the sake of existence and uniqueness of the solution to (\ref{stochastic_dynamics}).

Theorem~\ref{sic} analyzes stochastic incremental stability of two trajectories of (\ref{stochastic_dynamics}), $x_1$ and $x_2$. In Sec.~\ref{sec:cvstem}, we use it to find a contraction metric $M(x,t)$ for given $\alpha$, $\varepsilon$, and $L_m$, where $\alpha$ is a contraction rate, $\varepsilon$ is a parameter for disturbance attenuation, and $L_m$ is the Lipschitz constant of $M_{x_i}$. Note that $\varepsilon$ and $L_m$ are introduced for the sake of stochastic contraction and were not present in the deterministic case~\cite{ncm}. Sec.~\ref{given_params} delineates how we select them in practice.
\begin{theorem}
\label{sic}
Suppose $\exists g_1,g_2\in[0,\infty)$ \st{} $\|G(x_1,t)\|_F \leq g_1$ and $\|G(x_2,t)\|_F \leq g_2,~\forall x,t$. Suppose also that $\exists M(x,t) \succ 0$ \st{} $M_{x_i},~\forall x_i$ is Lipschitz with respect to the state $x$, \ie{} $\|M_{x_i}(x,t)-M_{x_i}(x',t)\|\leq L_m\|x-x'\|,~\forall x,x',t$ with $L_m \geq 0$. If $M(x,t)\succ 0$ and $\alpha,\varepsilon,\underline{\omega},\overline{\omega} \in(0,\infty)$ are given by
\begin{align}
\label{stochastic_contraction}
&\dot{M}(x,t)+2\sym\left(M(x,t)f_{x}(x,t)\right)+\alpha_gI\preceq -2\alpha M(x,t) \\
\label{Mcon}
& \overline{\omega}^{-1}I \preceq M(x,t) \preceq \underline{\omega}^{-1}I,~\forall x,t
\end{align}
where $\alpha_g = L_m(g_1^2+g_2^2)(\varepsilon+{1}/{2})$, then the mean squared distance between $x_1$ and $x_2$ is bounded as follows:
\begin{align}
\label{boundnewsto}
E\left[\|x_1-x_2\|^2\right] \leq \frac{C}{2\alpha}\frac{\overline{\omega}}{\underline{\omega}}+\overline{\omega}E[V(x(0),\delta x(0),0)]e^{-2\alpha t}.
\end{align}
where $V(x,\delta x,t) = \delta x^T M(x,t) \delta x$ and $C = (g_1^2+g_2^2)({2}/{\varepsilon}+1)$.
\end{theorem}
\begin{IEEEproof}
Let us first derive the bounds of $M_{x_i}$ and $M_{x_ix_j}$.
Since $M_{x_i},~\forall x_i$ is Lipschitz, we have $\|M_{x_i x_j}\| \leq L_m,~\forall i,j$ by definition. For $h\geq0$ and a unit vector $e_i$ with $1$ in its $i$th element, the Taylor's theorem suggests $\exists \xi_{-},\xi_{+}\in\mathbb{R}^n$ \st{}
\begin{align}
    M(x\pm he_i,t) = M(x,t)\pm M_{x_i}(x,t)h+M_{x_ix_i}(\xi_{\pm},t)h^2/2.
\end{align}
This implies that $\|M_{x_i}\|$ is bounded as $\left\|M_{x_i}\right\| \leq {h^{-1}\underline{\omega}^{-1}}+{L_mh}/{2} \leq \sqrt{{2L_m}{\underline{\omega}^{-1}}}$, where $h=\sqrt{2/(L_m\underline{\omega})}$ is substituted to obtain the last inequality.
Next, let $\mathscr{L}$ be the infinitesimal differential generator~\cite{mypaperTAC}. Computing $\mathscr{L}V$ using these bounds as in \cite{mypaperTAC} yields
\begin{align}
    \mathscr{L}V \leq& \delta x^T\left(\dot{M}+2\sym\left(Mf_{x}\right)\right)\delta x \nonumber \\
    &+(g_1^2+g_2^2)(L_m\|\delta x\|^2/2+2\sqrt{{2L_m}{\underline{\omega}^{-1}}}\|\delta x\|+\underline{\omega}^{-1}) \nonumber \\ 
    \leq& \delta x^T\left(\dot{M}+2\sym\left(Mf_{x}\right)+\alpha_gI\right)\delta x+{C}{\underline{\omega}^{-1}}
\end{align}
where the relation $2ab \leq \varepsilon^{-1}a^2+\varepsilon b^2$, which holds for any $a,b\in\mathbb{R}$ and $\varepsilon > 0$, is used with $a = \sqrt{2/\underline{\omega}}$ and $b = \sqrt{L_m}\|\delta z\|$ to get the second inequality. This reduces to $\mathscr{L}V \leq -2\alpha V+{C}\underline{\omega}^{-1}$ under the condition (\ref{stochastic_contraction}). The result (\ref{boundnewsto}) follows as in the proof of Theorem 1 in \cite{mypaperTAC}. 
\end{IEEEproof}
\begin{remark}
Note that there is a trade-off in using large $\varepsilon$ in Theorem~\ref{sic}, as it yields small $C$ to decrease the steady-state error in (\ref{boundnewsto}), but renders the constraint (\ref{stochastic_contraction}) tighter.
\end{remark}

Lemma~\ref{equiv_constraints} is used to convexify the cost function in Sec.~\ref{sec:cvstem}.
\begin{lemma}
\label{equiv_constraints}
The inequalities (\ref{stochastic_contraction}) and (\ref{Mcon}) are equivalent to
\begin{align}
\label{stochastic_contraction_tilde}
&\begin{bmatrix}
-\dot{\bar{W}}+2\sym{}(f_{x}(x,t)\bar{W})+2\alpha \bar{W} & \bar{W} \\ \bar{W} & -\frac{\nu}{\alpha_g}I
\end{bmatrix} \preceq 0
\\
\label{W_tilde}
&I \preceq \bar{W} \preceq \chi I,~\forall x,t
\end{align}
where $\nu = 1/\underline{\omega}$, $\chi = \overline{\omega}/\underline{\omega}$, and $\bar{W} = \nu W = \nu M^{-1}$.
\end{lemma}
\begin{IEEEproof}
Multiplying both sides of (\ref{stochastic_contraction}) by $W\succ 0$ and then by $\nu>0$ preserves matrix definiteness~\cite[pp. 114]{lmi}. This operation with Schur's complement lemma~\cite[pp. 28]{lmi} yield (\ref{stochastic_contraction_tilde}). The rest follows the proof of Lemma 1 of~\cite{ncm}.
\end{IEEEproof}
\begin{remark}
The variable conversion in Lemma~\ref{equiv_constraints} is necessary to get a convex cost function (\ref{cvstem_eq}) from the non-convex cost (\ref{boundnewsto}) as $t\to \infty$. 
In Sec.~\ref{sec:cvstem}, we use it to derive a semi-definite program in terms of $\nu$, $\chi$, and $\bar{W}$ for finding a contraction metric computationally efficiently~\cite{citeulike:163662}.
We show in Proposition~\ref{cvstem} that this is equivalent to the non-convex problem of minimizing (\ref{boundnewsto}) as $t\to \infty$, subject to (\ref{stochastic_contraction}) and (\ref{Mcon}) in terms of the original decision variables $\overline{\omega}$, $\underline{\omega}$, and $M$~\cite{mypaperTAC}.
\end{remark}

Finally, Lemma~\ref{sdclemma} introduces the generalized SDC form of dynamical systems to be exploited also in Sec.~\ref{sec:cvstem}.
\begin{lemma}
\label{sdclemma}
Suppose that $f(x,t)$ and $B(x,t)$ are continuously differentiable. Then $\exists A(x,x_d,t)$ \st{} $A(x,x_d,t)(x-x_d)=f(x,t)+B(x,t)u_d(x_d,t)-f(x_d,t)-B(x_d,t)u_d(x_d,t)$, $\forall x,x_d,u_d,t$, and one such $A$ is given as follows:
\begin{align}
\label{sdcAc}
A(x,x_d,t) =& \int_0^1\frac{\partial \bar{f}}{\partial x}(c x+(1-c)x_d,t)dc
\end{align}
where $\bar{f}(q,t)=f(q,t)+B(q,t)u_d(x_d,t)$. We call $A$ an SDC form when it is constructed to satisfy controllability and observability conditions (see Theorem~\ref{nscm_con} and Corollary~\ref{nscm_est}).
\end{lemma}
\begin{IEEEproof}
This follows from the integral relation given as $\int_0^1(d\bar{f}(c x+(1-c)x_d,t)/dc)dc=\bar{f}(x,t)-\bar{f}(x_d,t)$.
\end{IEEEproof}
\section{Neural Stochastic Contraction Metrics}
\label{sec:nscm}
 This section illustrates how to construct an NSCM using state samples $S=\{x_i\}_{i=1}^{N}$ and stochastic contraction metrics given by Theorem~\ref{sic}. This is analogous to the NCM~\cite{ncm}, which gives an optimal contraction metric for nonlinear systems with bounded disturbances, but the NSCM explicitly accounts for unbounded stochastic disturbances. For simplicity, we denote the metric both for feedback control and estimation as $X$ with $\underline{m}I \preceq X \preceq \overline{m}I$, \ie{}, $\underline{m}=\overline{\omega}^{-1}$, $\overline{m}=\underline{\omega}^{-1}$, $X=M$ for control, and $\underline{m}=\underline{\omega}$, $\overline{m}=\overline{\omega}$, $X=W$ for estimation.
\subsection{Data Pre-processing}
\label{data_preprocessing}
Since $X\succ0$, where $X$ is a contraction metric for control or estimation, it has a unique upper triangular matrix $Y \in\mathbb{R}^{n\times n}$ with positive diagonal entries \st{} $X = Y^TY$~\cite[pp. 441]{10.5555/2422911}. We use the nonzero entries of $Y$, denoted as $\theta (x,t) \in \mathbb{R}^{n(n+1)/2}$, for $y_i$ of (\ref{neuralnet}) to reduce its output dimension~\cite{ncm}.
\subsection{Lipschitz Condition and Spectral Normalization (SN)}
\label{sec_sn_nscm}
We utilize SN in Sec.~\ref{spectral} to guarantee the Lipschitz condition of Theorem~\ref{sic} or Proposition~\ref{cvstem} in Sec.~\ref{sec:cvstem}.
\begin{proposition}
\label{sn_proposition}
Let $\vartheta(x;W_{sn})$ be a neural network (\ref{neuralnet}) to model $\theta(x,t)$ in Sec.~\ref{data_preprocessing}, and $N_{\rm units}$ be the number of neurons in its last layer. Also, let $W_{sn}=\{W_{\ell}\}_{\ell=1}^{L+1}$, where $W_{\ell}=(\Omega_{\ell}/\|\Omega_{\ell}\|)C_{nn}$ for $1\leq\ell\leq L$, and $W_{\ell}=\sqrt{\overline{m}}(\Omega_{\ell}/\|\Omega_{\ell}\|)/\sqrt{N_{\rm units}}$ for $\ell=L+1$. If $\exists C_{nn},L_{m} > 0$ \st{}
\begin{align}
\label{sn_condition}
&2\left\|\vartheta_{x_i}(x;W_{sn})\right\|\left\|\vartheta_{x_j}(x;W_{sn})\right\| \\
&+2\|\vartheta(x;W_{sn})\|\|\vartheta_{x_i x_j}(x;W_{sn})\| \leq L_{m},~\forall i,j,x,\Omega \nonumber
\end{align}
then we have $\|\mathcal{X}\| \leq \overline{m}$ and $\|\mathcal{X}_{x_ix_j}\|\leq L_{m},~\forall x_i,x_j$, where $\mathcal{X}$ is the neural network model for the contraction metric $X(x,t)$. The latter inequality implies $\mathcal{X}_{x_i},~\forall i$ is indeed Lipschitz continuous with 2-norm Lipschitz constant $L_m$.
\end{proposition}
\begin{IEEEproof}
Let $\mathcal{Y}$ be the neural net model of $Y$ in Sec.~\ref{data_preprocessing}. By definition of $X=Y^TY$ and $\theta$, where $X$ is the contraction metric, we have $\|\mathcal{X}\|\leq\|\mathcal{Y}\|^2\leq\|\mathcal{Y}\|^2_F=\|\vartheta\|^2$. Thus, the relation $\|\vartheta(x;W_{sn})\|\leq \sqrt{N_{\rm units}}\|W_{L+1}\|$ yields $\|\mathcal{X}\|\leq {\overline{m}}$ for $W_{L+1}=\sqrt{\overline{m}}(\Omega_{L+1}/\|\Omega_{L+1}\|)/\sqrt{N_{\rm units}}$. Also, differentiating $\mathcal{X}$ twice yields $\|\mathcal{X}_{x_ix_j}\|/2 \leq \|\mathcal{Y}_{x_i}\|\|\mathcal{Y}_{x_j}\|+\|\mathcal{Y}\|\|\mathcal{Y}_{x_ix_j}\| \leq \|\vartheta_{x_i}\|\|\vartheta_{x_j}\|+\|\vartheta\|\|\vartheta_{x_ix_j}\|$, where the second inequality is due to $\|\mathcal{Y}\| \leq \|\mathcal{Y}\|_F = \|\vartheta\|$. Substituting $W_{sn}$ gives (\ref{sn_condition}). 
\end{IEEEproof}
\begin{example}
To see how Proposition~\ref{sn_proposition} works, let us consider a scalar input/output neural net with one neuron at each layer in (\ref{neuralnet}). Since we have~$\|\vartheta
(x;W_{sn})\| \leq \|W_{L+1}\|$, $\mathcal{X}\preceq \overline{m} I$ is indeed guaranteed by $\|W_{L+1}\| = \sqrt{\overline{m}}$. Also, we can get the bounds as $\|\vartheta_{x}(x;W_{sn})\| \leq \sqrt{\overline{m}}C_{nn}^L$ and $\|\vartheta_{xx}(x;W_{sn})\| \leq \|W_{L+1}\|C_{nn}^{L}(\sum_{\ell=1}^{L}C_{nn}^{\ell}) = \sqrt{\overline{m}}C_{nn}^{L+1}{(C_{nn}^L-1)}/{(C_{nn}-1)}$ using SN. Thus, (\ref{sn_condition}) can be solved for $C_{nn}$ by standard nonlinear equation solvers, treating $\overline{m}$ and $L_{m}$ as given constants.
\end{example}
\begin{remark}
\label{adaptive_remark}
For non-autonomous systems, we can treat $t$ or time-varying parameters $p(t)$ as another input to the neural network (\ref{neuralnet}) by sampling them in a given parameter range of interest. For example, we could use $p=[x_d,u_d]^T$ for systems with a target trajectory. This also allows us to use adaptive control techniques~\cite{Slotine:1228283,lopez2019contraction} to update an estimate of $p$.
\end{remark}
\section{mCV-STEM Sampling of Contraction Metrics} 
\label{sec:cvstem}
We introduce the modified ConVex optimization-based Steady-state Tracking
Error Minimization (mCV-STEM) method, an improved version of CV-STEM~\cite{mypaperTAC} for sampling the metrics which minimize an upper bound of the steady-state mean squared tracking error via convex optimization.
\begin{remark}
\label{combination_remark}
Due to its contraction-based formulation, combination properties~\cite{contraction} also apply to the NSCM framework. For example, contraction is preserved through hierarchical combination of estimation and control (\ie{} output feedback control), or through time-delayed communications~\cite{1618853}.
\end{remark}
\subsection{Stability of Generalized SDC Control and Estimation}
\label{sec:cvstemcontrol}
We utilize the general SDC parametrization with a target trajectory (\ref{sdcAc}), which captures nonlinearity through $A(x,x_d,t)$ or through multiple non-unique $A_i$~\cite{observer}, resulting in global exponential stability if the pair $(A,B)$ of (\ref{sdc_dynamics}) is uniformly controllable~\cite{sddre,observer}. Note that $x_d$ and $u_d$ can be regarded as extra inputs to the NSCM as in Remark~\ref{adaptive_remark}, but we could use Corollary~\ref{simple_sdc} as a simpler formulation which guarantees local exponential stability without using a target trajectory. Further extension to control contraction metrics, which use differential state feedback $\delta u = K(x,t) \delta x$~\cite{7852456,7989693,47710,WANG201944,vccm}, could be considered for sampling the metric with global reference-independent stability guarantees, achieving greater generality at the cost of added computation. Similarly, while we construct an estimator with global stability guarantees using the SDC form as in (\ref{estimator}),  a more general formulation could utilize geodesics distances between trajectories~\cite{scrm}. We remark that these trade-offs would also hold for deterministic control and estimation design via NCMs~\cite{ncm}.
\subsubsection{Generalized SDC Control}
Consider the following system with a controller $u \in \mathbb{R}^{m}$ and perturbation $\mathcal{W}(t)$:
\begin{align}
\label{sdc_dynamics}
dx =& (f(x,t)+B(x,t)u)dt+G_c(x,t)d\mathcal{W}(t) \\
\label{sdc_dynamicsd}
dx_d =& (f(x_d,t)+B(x_d,t)u_d(x_d,t))dt
\end{align}
where $B:\mathbb{R}^n\times\mathbb{R}_{\geq0}\to\mathbb{R}^{n\times m}$, $G_c:\mathbb{R}^n\times\mathbb{R}_{\geq0} \to \mathbb{R}^{n\times d}$, $\mathcal{W}(t)$ is a $d$-dimensional Wiener process, and $x_d:\mathbb{R}_{\geq0} \to \mathbb{R}^{n}$ and $u_d:\mathbb{R}^n\times\mathbb{R}_{\geq0} \to \mathbb{R}^{m}$ denote the target trajectory. 
\begin{theorem}
\label{nscm_con}
Suppose $\exists g_c \in [0,\infty)$ \st{} $\|G_c(x,t)\|_F\leq g_c,~\forall x,t$, and $\exists M(x,x_d,t)\succ0$ \st{} $M_{x_i}$ and $M_{x_{d,i}},~\forall x_i,x_{d,i}$ are Lipschitz with respect to its state with 2-norm Lipschitz constant $L_m$. Let $u$ be designed~as
\begin{align}
\label{controller}
&u = u_d(x_d,t)-B(x,t)^TM(x,x_d,t)(x-x_d) \\
\label{controller_con}
&\dot{M}+2\sym(MA)-2MBB^TM+\alpha_{gc}I \preceq -2\alpha M \\
\label{controller_con2}
&\overline{\omega}^{-1}I \preceq M(x,x_d,t) \preceq \underline{\omega}^{-1}I,~\forall x,t
\end{align}
where $\alpha>0$, $\alpha_{gc} = L_mg_c^2(\varepsilon+{1}/{2})$, $\varepsilon>0$, and $A$ is given by (\ref{sdcAc}) in Lemma~\ref{sdclemma}. If the pair $(A,B)$ is uniformly controllable, we have the following bound for the systems (\ref{sdc_dynamics}) and (\ref{sdc_dynamicsd}):
\begin{align}
\label{control_ss}
E[\|x-x_d\|^2] \leq \frac{C_c}{2\alpha}\chi+\overline{\omega}E[V(x(0),x_d(0),\delta q(0),0)]e^{-2\alpha t}
\end{align}
where $V(x,x_d,\delta q,t)=\delta q^TM(x,x_d,t)\delta q$, $C_c = g_c^2(2/\varepsilon+1)$, $\nu = {1}/{\underline{\omega}}$, $\chi = {\overline{\omega}}/{\underline{\omega}}$, and $q$ is the state of the differential system with its particular solutions $q=x,x_d$.
Further, (\ref{controller_con}) and (\ref{controller_con2}) are equivalent to the following constraints in terms of $\nu$, $\chi$, and $\bar{W} = \nu W=\nu M^{-1}$:
\begin{align}
\label{convex_constraint_controller1}
&\begin{bmatrix}
-\dot{\bar{W}}+2\sym{}(A\bar{W})-2\nu BB^T+2\alpha \bar{W} & \bar{W} \\ \bar{W} & -\frac{\nu}{\alpha_{gc}}I
\end{bmatrix} \preceq 0 \\
\label{convex_constraint_controller2}
&I\preceq \bar{W}\preceq \chi I,~\forall x,t.
\end{align}
where the arguments are omitted for notational simplicity.
\end{theorem}
\begin{IEEEproof}
Using the SDC parameterization (\ref{sdcAc}) given in Lemma~\ref{sdclemma}, (\ref{sdc_dynamics}) can be written as $d{x} = (\bar{f}(x_d,t)+(A(x,x_d,t)-B(x,t)B(x,t)^TM(x,x_d,t))(x-x_d))dt+G_c(x,t)d\mathcal{W}$. This results in the following differential system, $dq = (\bar{f}(x_d,t)+(A(x,x_d,t)-B(x,t)B(x,t)^TM)(q-x_d))dt+G(q,t)d\mathcal{W}$, where $G(q,t)$ is defined as $G(q=x,t)=G_c(x,t)$ and $G(q=x_d,t)=0$. Note that it has $q=x,x_d$ as its particular solutions. Since $f_x$, $g_1$, and $g_2$ in Theorem~\ref{sic} can be viewed as $A(x,x_d,t)-B(x,t)B(x,t)^TM(x,x_d,t)$, $g_c$, and $0$, respectively, applying its results for $V = \delta q^T M(x,x_d,t) \delta q$ gives (\ref{control_ss}) as in (\ref{boundnewsto}). The constraints (\ref{convex_constraint_controller1}) and (\ref{convex_constraint_controller2}) follow from the application of Lemma~\ref{equiv_constraints} to (\ref{controller_con}) and (\ref{controller_con2}).
\end{IEEEproof}
\begin{remark}
For input non-affine nonlinear systems, we can find $f(x,u)-f(x_d,u_d)=A(x,u,t)(x-x_d)+B(x,u,t)(u-u_d)$ by Lemma~\ref{sdclemma} and use it in Theorem~\ref{nscm_con}, although (\ref{controller}) has to be solved implicitly as $B$ depends on $u$ in this case~\cite{vccm,WANG201944}. 
\end{remark}
\subsubsection{Generalized SDC Estimation}
Consider the following system and a measurement $y(t)$ with perturbation $\mathcal{W}_{1,2}(t)$:
\begin{align}
\label{sdc_dynamics_est}
dx =& f(x,t)dt+G_e(x,t)d\mathcal{W}_1(t) \\
\label{sdc_measurement}
ydt =& h(x,t)dt+D(x,t)d\mathcal{W}_2(t)
\end{align}
where $h:\mathbb{R}^n\times\mathbb{R}_{\geq0} \to \mathbb{R}^{m}$, $G_e:\mathbb{R}^n\times\mathbb{R}_{\geq0} \to \mathbb{R}^{n\times d_1}$, $D:\mathbb{R}^n\times\mathbb{R}_{\geq0} \to \mathbb{R}^{m\times d_2}$, and $\mathcal{W}_{1,2}(t)$ are two independent Wiener processes.
We have an analogous result to Theorem~\ref{nscm_con}. 
\begin{corollary}
\label{nscm_est}
Suppose $\exists g_e,\overline{d}\in[0,\infty)$ \st{} $\|G_e(x,t)\|_F\leq g_e$ and $\|D(x,t)\|_F\leq \overline{d},~\forall x,t$. Suppose also that $\exists W(\hat{x},t)=M(\hat{x},t)^{-1}\succ 0$ \st{} $W_{x_i},~\forall x_i$ is Lipschitz with respect to its state with 2-norm Lipschitz constant $L_m$. Let $\nu=1/\underline{\omega}$ and $x$ be estimated as
\begin{align}
\label{estimator}
&d\hat{x} = f(\hat{x},t)dt+M(\hat{x},t)C_L(\hat{x},t)^T(y-h(\hat{x},t))dt \\
\label{estimator_con1}
&\dot{W}+2\sym(W A-C^T_LC)+\alpha_{ge}I
 \preceq -2\alpha W \\
\label{estimator_con2}
&\underline{\omega}I \preceq W(\hat{x},t) \preceq \overline{\omega}I,~0<\nu\leq\sqrt[3]{\nu_{c}},~\forall x,\hat{x},t
\end{align}
where $\alpha,\nu_c,\varepsilon >0$, $\alpha_{ge} = \alpha_{e1}+\nu_c\underline{\omega}\alpha_{e2}$, $\alpha_{e1} = L_mg_e^2(\varepsilon+{1}/{2})$, and $\alpha_{e2} = L_m\overline{c}^2\overline{d}^2(\varepsilon+{1}/{2})$. Also, $A(x,\hat{x},t)$ and $C(x,\hat{x},t)$ are given by (\ref{sdcAc}) of Lemma~\ref{sdclemma} with $(f,x,x_d,u_d)$ replaced by $(f,\hat{x},x,0)$ and $(h,\hat{x},x,0)$, respectively, and $C_L(\hat{x},t)=C(\hat{x},\hat{x},t)$.
If $(A,C)$ is uniformly observable and
$\|C(x,\hat{x},t)\|\leq\overline{c},~\forall x,\hat{x},t$, then we have the following bound:
\begin{align}
\label{estimator_ss}
E[\|x-\hat{x}\|^2] \leq \frac{C_e}{2\alpha}+\frac{1}{\underline{\omega}}E[V(x(0),\delta q(0),0)]e^{-2\alpha t}
\end{align}
where $V(\hat{x},\delta q,t)=\delta q^T W(\hat{x},t)\delta q$, $C_e=C_{e1}\chi+C_{e2}\chi\nu^2$, $C_{e1} = g_e^2(2/\varepsilon+1)$, $C_{e2} = \overline{c}^2\overline{d}^2(2/\varepsilon+1)$, $\chi = {\overline{\omega}}/{\underline{\omega}}$, and $q$ is the state of the differential system with its particular solutions $q=\hat{x},x$. Further, (\ref{estimator_con1}) and (\ref{estimator_con2}) are equivalent to the following constraints in terms of $\nu$, $\nu_c$, $\chi$, and $\bar{W} = \nu W$:
\begin{align}
    \label{convex_constraint_estimator1}
    &\dot{\bar{W}}+2\sym{}(\bar{W}A-\nu C^T_LC)+\nu\alpha_{e1}I +\nu_c\alpha_{e2}I \preceq -2\alpha \bar{W} \\
    \label{convex_constraint_estimator2}
    &I\preceq \bar{W}\preceq \chi I,~0<\nu\leq\sqrt[3]{\nu_{c}},~\forall x,\hat{x},t
\end{align}
where the arguments are omitted for notational simplicity.
\end{corollary}
\begin{IEEEproof}
The differential system of (\ref{sdc_dynamics_est}) and (\ref{estimator}) is given as $dq = f(x,t)+(A(x,\hat{x},t)-M(\hat{x},t)C_L(\hat{x},t)^TC(x,\hat{x},t))(q-x))dt+G(q,t)d\mathcal{W}$, where $G(q,t)$ is defined as $G(q=x,t)=G_e(x,t)$ and $G(q=\hat{x},t)=M(\hat{x},t)C(\hat{x},t)^TD(x,t)$. Viewing $V$, $g_1$, and $g_2$ in Theorem~\ref{sic} as $V = \delta q^T W(\hat{x},t) \delta q$, $g_1 = g_e$, and $g_2 = \overline{c}\overline{d}/\underline{\omega}$, (\ref{estimator_ss}) -- (\ref{convex_constraint_estimator2}) follow as in the proof of Theorem~\ref{nscm_con} due to $\nu^3=\underline{\omega}^{-3}\leq\nu_c$ and the contraction condition (\ref{estimator_con1}).
\end{IEEEproof} 

Note that (\ref{controller_con}) and (\ref{estimator_con1}) depend on their target trajectory, \ie{}, $x_d$ for control and $x$ for estimation. We can treat them as time-varying parameters $p(t)$ in a given space during the mCV-STEM sampling as in Remark~\ref{adaptive_remark}. Alternatively, we could use the following to avoid this complication.
\begin{corollary}
\label{simple_sdc}
Using predefined trajectories (\eg{} $(x_d,u_d)=(0,0)$ for control or $x=0$ for estimation) in Thm.~\ref{nscm_con} or Cor.~\ref{nscm_est} leads to local exponential stability of (\ref{sdc_dynamics}) or (\ref{estimator}).
\end{corollary}
\begin{IEEEproof}
This follows as in the proof of Thm.~\ref{nscm_con}~\cite{ncm}.
\end{IEEEproof}
\subsection{mCV-STEM Formulation}
The following proposition summarizes the mCV-STEM.
\begin{proposition}
\label{cvstem}
The optimal contraction metric $M = W^{-1}$ that minimizes the upper bound of the steady-state mean squared distance ((\ref{control_ss}) of Thm.~\ref{nscm_con} or (\ref{estimator_ss}) of Corr.~\ref{nscm_est} with $t\rightarrow \infty $) of stochastically perturbed system trajectories is found by the following convex optimization problem:
\begin{align}
\label{cvstem_eq}
&{J}_{CV}^* = \min_{\nu>0,\nu_c>0,\chi \in \mathbb{R},\bar{W} \succ 0} c_1\chi+c_2\nu+c_3 P(\nu,\nu_c,\chi,\bar{W}) \\
&\text{\st{}~~(\ref{convex_constraint_controller1}) \& (\ref{convex_constraint_controller2}) for control, (\ref{convex_constraint_estimator1}) \& (\ref{convex_constraint_estimator2}) for estimation} \nonumber
\end{align}
where $c_1,c_2,c_3\in[0,\infty)$ and $P$ is an additional performance-based convex cost (see Sec.~\ref{sec:ChoiceObJ}). The weight of $\nu$, $c_2$, can either be viewed as a penalty on the 2-norm of feedback gains or an indicator of how much we trust the measurement $y(t)$. Note that $\alpha$, $\varepsilon$, and $L_m$ are assumed to be given in (\ref{cvstem_eq}) (see Sec.~\ref{given_params} for how to handle $\dot{\bar{W}}$ preserving convexity).
\end{proposition}
\begin{IEEEproof}
For control (\ref{control_ss}), using $c_1=C_c/(2\alpha)$ and $c_2=c_3=0$ gives (\ref{cvstem_eq}). We can set $c_2 > 0$ to penalize excessively large $\|u\|$ through $\nu \geq \sup_{x,t}\|M(x,x_d,t)\|$. Since we have $\nu>0$ and $1 \leq \chi \leq \chi^3$, (\ref{estimator_ss}) as $t\to \infty$ can be bounded as
\begin{align}
\label{est_bound_example}
\frac{C_{e1}\chi+C_{e2}\chi\nu^2}{2\gamma} \leq \frac{1}{3\sqrt{3C_{e1}}}\left(\frac{\sqrt{3C_{e1}}}{\sqrt[3]{2\gamma}}\chi+\frac{\sqrt{C_{e2}}}{\sqrt[3]{2\gamma}}\nu\right)^3.
\end{align}
Minimizing the right-hand side of (\ref{est_bound_example}) gives (\ref{cvstem_eq}) with $c_1=\sqrt{3C_{e1}}/\sqrt[3]{2\gamma}$, $c_2=\sqrt{C_{e2}}/\sqrt[3]{2\gamma}$, and $c_3=0$. Finally, since $\overline{d}=0$ in (\ref{sdc_measurement}) means $C_{e2}=0$ and no noise acts on $y$,~$c_2$ also indicates how much we trust the measurement.
\end{IEEEproof}
\subsubsection{Choice of $P(\nu,\nu_c,\chi,\bar{W})$}\label{sec:ChoiceObJ}
Selecting $c_3=0$ in Proposition~\ref{cvstem} yields an affine objective function which leads to a straightforward interpretation of its weights. Users could also select $c_3>0$ with other performance-based cost functions $P(\nu,\nu_c,\chi,\bar{W})$ in (\ref{cvstem_eq}) as long as they are convex. For example, an objective function $\sum_{x_i\in S}\|u\|^2=\sum_{x_i\in S}\|-B(x_i,t)^TM(x_i,t)x_i\|^2\leq\sum_{x_i\in S}\|B(x_i,t)\|^2\|x_i\|^2\nu^2$, where $S$ is the state space of interest, gives an optimal contraction metric which minimizes the upper bound of its control effort.
\subsubsection{Additional Parameters and $\dot{\bar{W}}$}
\label{given_params}
We assumed $\alpha$, $\varepsilon$, and $L_m$ are given in Proposition~\ref{cvstem}. For $\alpha$ and $\varepsilon$, we perform a line search to find their optimal values as will be demonstrated in Sec.~\ref{simulation}. For $L_m$, we guess it by a deterministic NCM~\cite{ncm} and guarantee the Lipschitz condition by SN as explained in Sec.~\ref{sec_sn_nscm}. Also, (\ref{cvstem_eq}) can be solved as a finite-dimensional problem by using backward difference approximation on $\dot{\bar{W}}$, where we can then use $-\bar{W} \preceq -I$ to obtain a sufficient condition of its constraints, or solve it along pre-computed trajectories $\{x(t_i)\}_{i=0}^M$~\cite{ncm,doi:10.1162/neco.1997.9.8.1735}.
The pseudocode to obtain the NSCM depicted in Fig.~\ref{nscmdrawing} is given in Algorithm~\ref{alg}.
\begin{algorithm}
\SetKwInOut{Input}{Inputs}\SetKwInOut{Output}{Outputs}
\Input{States \& parameters: $S = \{x_{i}\}^{N}_{i=1}$ or $\{\hat{x}_{i}\}^{N}_{i=1}$ \& $T = \{p_i\}_{i=1}^{M}$ (\eg{} $p=t$, $[x_d,u_d]^T$, or $x$)}
\Output{NSCM and $J_{CV}^*$ in (\ref{cvstem_eq})}
\BlankLine
\textit{1. Sampling of Optimal Contraction Metrics} \\
Find $L_m$ in Thm.~\ref{sic} using a deterministic NCM~\cite{ncm} \\
\For{$(\alpha,\varepsilon) \in A_{LS}\text{ ($A_{LS}$ is a search set )}$}{
Solve (\ref{cvstem_eq}) of Prop.~\ref{cvstem} using $x,p$ or $\hat{x},p$ in $S$ \& $T$ \\
Save the optimizer $(\nu,\chi,\{\bar{W}_i\}_{i=1}^{N})$ and optimal value $J(\alpha,\varepsilon) = c_1\chi+c_2\nu+c_3P(\nu,\nu_c,\chi,\bar{W})$
} 
Find $(\alpha^*,\varepsilon^*) = \text{arg}\min_{(\alpha,\varepsilon) \in A_{LS}}J$ and $J_{CV}^*=J(\alpha^*,\varepsilon^*)$\\
Obtain $(\nu(\alpha^*,\varepsilon^*),\chi(\alpha^*,\varepsilon^*),\{\bar{W}_i(\alpha^*,\varepsilon^*)\}_{i=1}^{N})$ 
\BlankLine
\textit{2. Spectrally-Normalized Neural Network Training} \\
Pre-process data as in Sec.~\ref{data_preprocessing} \\
Split data into a train set $\mathcal{S}_{\mathrm{train}}$ and test set $\mathcal{S}_{\mathrm{test}}$ \\
\For{$\mathrm{epoch} \leftarrow 1$ \KwTo $N_{\mathrm{epochs}}$}{
\For{$s \in \mathcal{S}_{\mathrm{train}}$}{Train a neural network using SGD with the Lipschitz condition on $X_{x_i}$ as in Prop.~\ref{sn_proposition}}
Compute the test error for data in $\mathcal{S}_{\mathrm{test}}$ \\
\If{\rm test error is small enough}{
\textbf{break}
}
}
\caption{NSCM Algorithm}
\label{alg}
\end{algorithm}
\section{Numerical Implementation Example}
\label{simulation}
We demonstrate the NSCM on a rocket autopilot problem ({\color{caltechgreen}\underline{\href{https://github.com/astrohiro/nscm}{https://github.com/astrohiro/nscm}}}). CVXPY~\cite{cvxpy} with the MOSEK solver~\cite{mosek} is used to solve convex optimization.
\subsection{Simulation Setup}
\begin{figure}
    \centering
    \includegraphics[width=35mm]{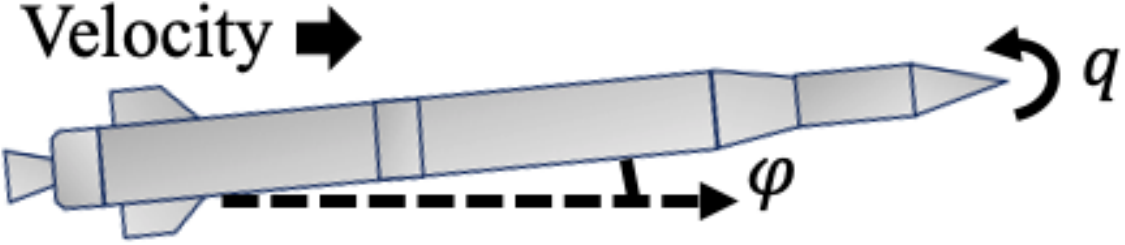}
    \caption{Rocket model (angle of attack $\varphi$, pitch rate $q$).}
    \label{rocket_model}
    \vspace{-0.7em}
\end{figure}

We use the nonlinear rocket model in Fig.~\ref{rocket_model}~\cite{doi:10.2514/3.20997}, assuming $q$ and specific normal force are available via rate gyros and accelerometers. We use $G_c=(6.0$e--$2)I_{n}$, $G_e=(3.0$e--$2)I_n$, and $D=(3.0$e--$2)I_m$ for perturbation in the NSCM construction. The Mach number is varied linearly in time from $2$ to $4$.
\subsection{NSCM Construction}
\begin{figure} 
    \centering
    \includegraphics[width=50mm]{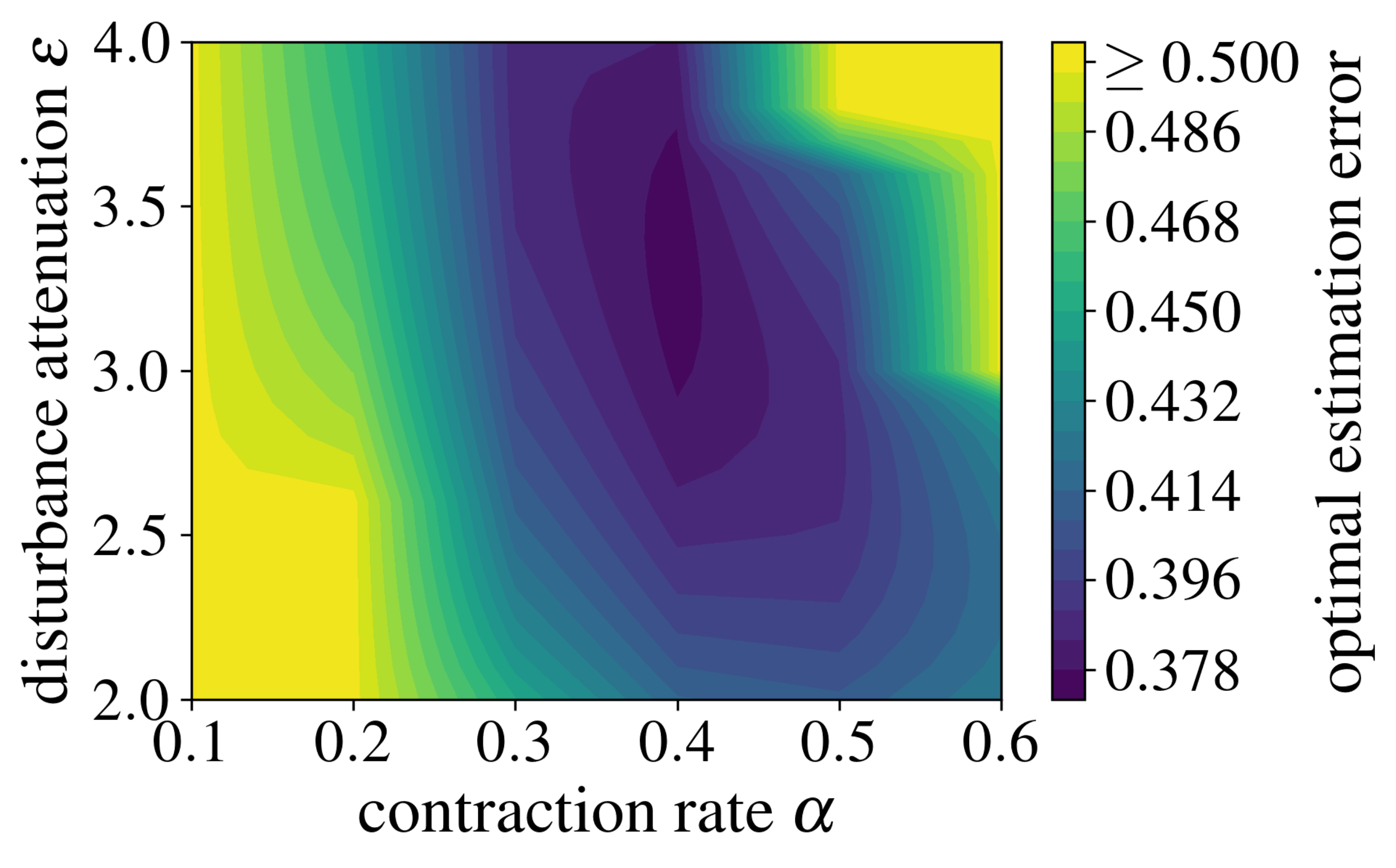}
    \vspace{-1.2em}
    \caption{Optimal steady-state estimation error as a function of $\alpha$ and $\varepsilon$.}
    \label{alpeps}
    \vspace{-0.8em}
\end{figure}

\begin{figure}
    \centering
    \includegraphics[width=70mm]{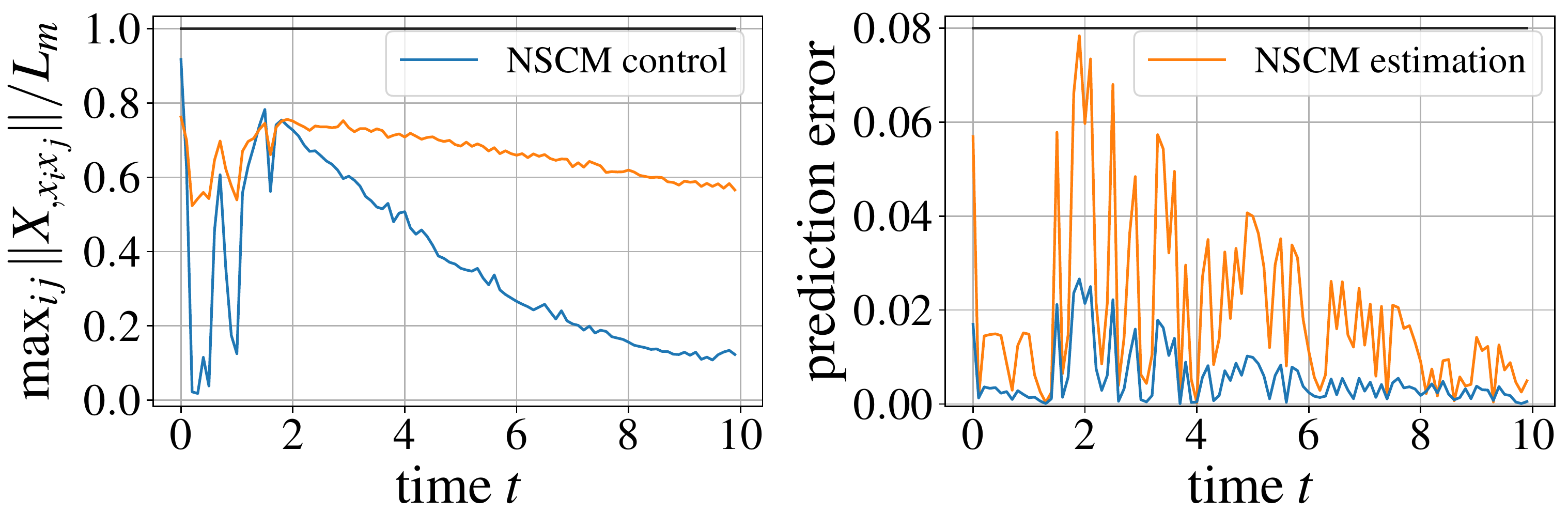}
    \vspace{-1.2em}
    \caption{NSCM spectral normalization and prediction error.}
    \label{nn_errors}
    \vspace{-1.0em}
\end{figure}

We construct NSCMs by Algorithm~\ref{alg}. For estimation, we select the Lipschitz constant on $X_{x_i}$ to be $L_m=0.50$ (see Sec.~\ref{given_params}). The optimal $\alpha$ and $\varepsilon$, $\alpha^* = 0.40$ and $\varepsilon^* = 3.30$, are found by line search in Fig.~\ref{alpeps}. A neural net with $3$ layers and $100$ neurons is trained using $N=1000$ samples, where its SN constant is selected as $C_{nn} = 0.85$ as a result of Proposition~\ref{sn_proposition}. We use the same approach for the NSCM control and the resultant design parameters are given in Table~\ref{optimal_est_con_values}. Figure~\ref{nn_errors} implies that the NSCMs indeed satisfy the Lipschitz condition with its prediction error smaller than $0.08$ thanks to SN.
\begin{table}
\vspace{-1.5em}
\caption{NSCM Control and Estimation Parameters}
\label{optimal_est_con_values}
\centering
\begin{tabular}{|c|c|c|c|c|} 
 \hline
   & $\alpha$ & $\varepsilon$ & $L_m$ & steady-state upper bound \rule[0mm]{0mm}{2.7mm} \\
 \hline
 estimation & $0.40$ & $3.30$ & $0.50$ & $0.41$ \rule[0mm]{0mm}{2.7mm} \\
 \hline
 control & $0.10$ & $1.00$ & $10.00$ & $0.58$ \rule[0mm]{0mm}{2.7mm} \\
 \hline
\end{tabular}
\end{table}
\begin{figure}
\vspace{-0.7em}
    \centering
    \includegraphics[width=80mm]{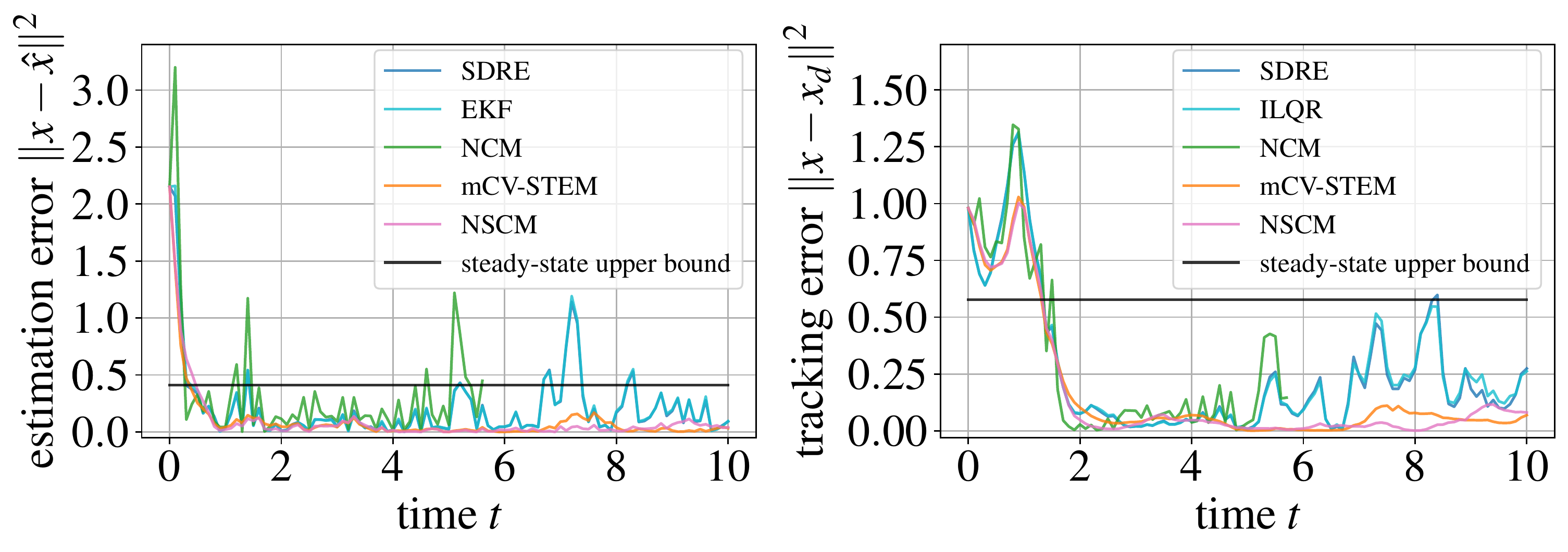}
    \vspace{-1.2em}
    \caption{Rocket state estimation and tracking error ($x = [\varphi,q]^T$).}
    \label{est_con_sim}
\end{figure}
\section{Discussion and Concluding Remarks}
We compare the NSCM with the SDRE~\cite{sddre}, ILQR~\cite{ilqr,silqr}, EKF, NCM~\cite{ncm}, and mCV-STEM. As shown in Fig.~\ref{est_con_sim}, the steady-state errors of the NSCM and mCV-STEM are indeed smaller than its steady-state upper bounds (\ref{control_ss}) and (\ref{estimator_ss}) found by Proposition~\ref{cvstem}, while other controllers violate this condition. 
Also, the optimal contraction rate of the NCM for state estimation is much larger ($\alpha = 6.1$) than the NSCM as it does not account for stochastic perturbation. This renders the NCM trajectory diverge around $t=5.8$ in Fig.~\ref{est_con_sim}. The NSCM Lipschitz condition on $X_{x_i}$ guaranteed by SN as in Fig.~\ref{nn_errors} allows us to circumvent this difficulty.

In conclusion, the NSCM is a novel way of using spectrally-normalized deep neural networks for real-time computation of approximate nonlinear control and estimation policies, which are optimal and provably stable in the mean squared error sense even under stochastic disturbances. We remark that the reference-independent policies~\cite{7852456,7989693,47710,WANG201944,vccm,scrm} or the generalized SDC policies (\ref{controller}) and (\ref{estimator}) introduced in this paper, which guarantee global exponential stability with respect to a target trajectory, could be used both in stochastic and deterministic frameworks including the NCM~\cite{ncm}. It is also noted that the combination properties of contraction theory in Remark~\ref{combination_remark} still holds for the deterministic NCM. An important future direction is to consider a model-free version of~these~techniques~\cite{boffi2020learning}.
\subsubsection*{Acknowledgments}
This work was funded in part by the Raytheon Company and benefited from discussions with Nicholas Boffi and Quang-Cuong Pham.
\bibliographystyle{IEEEtran}
\bibliography{ms}
\end{document}